\documentclass[11pt,a4paper]{article}
\usepackage[hyperref]{acl2019}
\usepackage{times}
\usepackage{latexsym}

\usepackage{times}
\usepackage{soul}
\usepackage{url}
\usepackage[utf8]{inputenc}
\usepackage{caption}
\usepackage{graphicx}
\usepackage{amsmath}
\usepackage{booktabs}
\usepackage{algorithm}
\usepackage{algorithmic}
\usepackage{times}
\usepackage{latexsym}
\usepackage{graphicx}
\usepackage{multirow}
\usepackage{url,paralist}
\usepackage{amsmath,amsbsy,amsthm,amssymb,bm,color,xcolor}
\usepackage[inline]{enumitem}

\newcommand{\intd}{\operatorname{d}}
\newcommand{\zsyn}{\bm z_\text{syn}}
\newcommand{\zsem}{\bm z_\text{sem}}

\newcommand{\method}{DSS-VAE }
\newcommand{\methodp}{DSS-VAE}
\DeclareMathOperator*{\E}{\mathbb{E}}
\DeclareMathOperator{\KL}{\operatorname{KL}}

\newcommand{\refsem}{\text{Ref}_\text{sem}}

\newcommand{\refsyn}{\text{Ref}_\text{syn}}

\makeatletter
\newcommand\incircbin
{%
  \mathpalette\@incircbin
}
\newcommand\@incircbin[2]
{%
  \mathbin%
  {%
    \ooalign{\hidewidth$#1#2$\hidewidth\crcr$#1\bigcirc$}%
  }%
}
\aclfinalcopy

\title{Generating Sentences from Disentangled Syntactic and Semantic Spaces}

\author{Yu Bao$^{1}$\footnotemark[1] \quad Hao Zhou$^{2}$\footnotemark[1] \quad Shujian Huang$^{1}$\footnotemark[2] \quad Lei Li$^{2}$ \quad Lili Mou$^{3}$ \\
\textbf{ Olga Vechtomova$^{3}$\quad Xinyu Dai$^{1}$\quad Jiajun Chen$^{1}$}\\
$^{1}$National Key Laboratory for Novel Software Technology, Nanjing University, China\\
$^{2}$ByteDance AI Lab, Beijing, China \\
$^{3}$University of Waterloo, Canada \\
\texttt{\{baoy,huangsj,dxy,chenjj\}@nlp.nju.edu.cn}\ \\ 
\texttt{\{zhouhao.nlp,lileilab\}@bytedance.com}\\ \texttt{doublepower.mou@gmail.com},\ \ \texttt{ovechtomova@uwaterloo.ca}
}

\begin{document}
\maketitle
\renewcommand{\thefootnote}{\fnsymbol{footnote}}
\begin{abstract}
Variational auto-encoders~(VAEs) are widely used in natural language generation due to the regularization of the latent space. 
However, generating sentences from the continuous latent space does not explicitly model the syntactic information.
In this paper, we propose to generate sentences from disentangled syntactic and semantic spaces.
Our proposed method explicitly models syntactic information in the VAE's latent space by using the linearized tree sequence, leading to better performance of language generation.
Additionally, the advantage of sampling in the disentangled syntactic and semantic latent spaces enables us to perform novel applications, such as the unsupervised paraphrase generation and syntax-transfer generation.  Experimental results show that our proposed model
achieves similar or better performance in various tasks, compared with state-of-the-art related work.
\end{abstract}

\footnotetext[1]{Equal contributions.}
\footnotetext[2]{Corresponding author.}
\renewcommand{\thefootnote}{\arabic{footnote}}
\setcounter{footnote}{0}
\section{Introduction}

Variational auto-encoders~\cite[VAEs,][]{vae} are widely used in language generation tasks~\cite{serban2017hierarchical,gvae,conv_vae,li2018generating}. 
VAE encodes a sentence into a probabilistic latent space, from which it learns to decode the same sentence. 
In addition to traditional reconstruction loss of an autoencoder, VAE employs an extra regularization term, penalizing the Kullback--Leibler~(KL) divergence between the encoded posterior distribution and its prior.
This property enables us to sample and generate sentences from the continuous latent space.
Additionally, we can even manually manipulate the latent space, inspiring various applications such as sentence interpolation~\cite{bowman} and text style transfer~\cite{hu}.

However, the continuous latent space of VAE blends syntactic and semantic information together, without modeling the syntax explicitly.
We argue that it may be not necessarily the best in the text generation scenario.
Recently, researchers have shown that explicitly syntactic modeling improves the generation quality in sequence-to-sequence models~\cite{eriguchi2016tree,zhou2017chunk,li2017modeling,chen2017improved}.
It is straightforward to adopt such idea in the VAE setting, since a vanilla VAE does not explicitly model the syntax. 
A line of studies~\cite{gvae,cvae,syntax_directed_vae} propose to impose context-free grammars~(CFGs) as hard constraints in the VAE decoder, so that they could generate syntactically valid outputs of programs, molecules, etc. 

However, the above approaches cannot be applied to syntactic modeling in VAE's continuous latent space, and thus, we do not enjoy the two benefits of VAE, namely, \textit{sampling} and \textit{manipulation}, towards the syntax of a sentence.

In this paper, we propose to generate sentences from a disentangled syntactic and semantic spaces of VAE~(called \methodp).
\method explicitly models syntax in the continuous latent space of VAE, while retaining the sampling and manipulation benefits.
In particular, we introduce two continuous latent variables to capture semantics and syntax, respectively. 
To separate the semantic and syntactic information from each other, we borrow the adversarial approaches from the text style-transfer research~\cite{hu,fu,vineet}, but adapt it into our scenario of syntactic modeling.
We also observe that syntax and semantics are highly interwoven, and therefore further propose an adversarial reconstruction loss to regularize the syntactic and semantic spaces. 

Our proposed \method takes following advantages:

First, explicitly syntactic modeling in VAE's latent space improves the quality of unconditional language generation. 
Experiments show that, compared with traditional VAE, \method generates more fluent sentences~(lower perplexity), while preserving more amount of encoded information~(higher BLEU scores for reconstruction).
Comparisons with a state-of-the-art syntactic language model~\cite{shen2017neural} are also included.

Second, the advantage of manipulation in the syntactic and semantic spaces of \method provides a natural way of unsupervised paraphrase generation.
If we sample a vector in the syntactic space but perform max \textit{a posterior}~(MAP) inference in the semantic space, we are able to generate a sentence with the same meaning but different syntax. 
This is known as \textit{unsupervised paraphrase generation}, as no parallel corpus is needed during training. 
Experiments show that \method outperforms the traditional VAE as well as a state-of-the-art Metropolis-Hastings sampling approach~\cite{MH} in this task.

Additionally, with the disentangled syntactic and semantic latent spaces, we propose an interesting application that transfers the syntax of one sentence to another.
Both qualitative and quantitative experimental results show that \method could graft the designed syntax to another sentence under certain circumstances.

\section{Related Work}

The variational auto-encoders~(VAEs) is proposed by \newcite{vae} for image generation. \newcite{bowman} successfully applied VAE in the NLP domain, showing that VAE improves recurrent neural network (RNN)-based language modeling~\cite[RNN-LM,][]{mikolov2010recurrent}; that VAE allows sentence sampling and sentence interpolation in the continuous latent space. Later, VAE is widely used in various natural language generation tasks~\cite{vae_svg_eq,gvae,hu,syntactic_mani}. 

Syntactic language modeling, to the best of our knowledge, could be dated back to~\newcite{chelba1997}. \newcite{charniak2001immediate} and \newcite{clark2001unsupervised} propose to utilize a top-down parsing mechanism for language modeling. \newcite{rnng} and \newcite{whatrnng} introduce the neural network to this direction. The Parsing-Reading-Predict Network~\cite[PRPN,][]{shen2017neural}, which reports a state-of-the-art results on syntactic language modeling, learns a latent syntax by training with a language modeling objective. Different from their work, our approach models syntax in a continuous space, facilitating sampling and manipulation of syntax.


Our work is also related to style-transfer text generation~\cite{fu,percy,vineet}. In previous work, the style is usually defined by categorical features such as sentiment. 
We move one step forward, extending their approach to the sequence level and dealing with more complicated, non-categorical syntactic spaces.
Due to the complication of syntax, we further design adversarial reconstruction losses to encourage the separation of syntax and semantics. 

\section{Approach}

In this section, we present our proposed \method in detail. We first introduce the variational autoencoder in \S\ref{ss:vanillaVAE}. Then, we describe the general architecture of \method in \S\ref{ss:syntaxVAE}, where we explain how we generate sentences from disentangled syntactic and semantic latent spaces and how we disentangle information from the two separated spaces. Model training is discussed in \S\ref{ss:train}.

\subsection{Variational Autoencoder}\label{ss:vanillaVAE}
A traditional VAE employs a  probabilistic latent variable $\bm z$ to encode the information of a sentence $\bm x$, and then decodes the original $\bm x$ from $\bm z$. 
The probability of a sentence $\bm x$ could be computed as: 
\begin{equation}
    p(\bm x)= \int p(\bm z) p(\bm x|\bm z)\intd\!\bm z
     \label{eqn:vae_prob}
\end{equation}
where $p(\bm z)$ is the prior, and $p(\bm x|\bm z)$ is given by the decoder. VAE is trained by maximizing the \textit{evidence lower bound}~(ELBO):
\begin{equation}\label{eqn:elbo}
    \begin{split}
        &\log p(\bm x) \ge \operatorname{ELBO}\\
   & = \E\limits_{q(\bm z |\bm x)}\big[\log p(\bm x|\bm z)\big]- \KL\left(q(\bm z|\bm x)\; \big\| \; p(\bm z)\right)
\end{split}
\end{equation}

\subsection{Proposed Method: \methodp}\label{ss:syntaxVAE}
Our \method is built upon the vanilla VAE, but extends Eqn.~(\ref{eqn:vae_prob}) by adopting two separate latent variables $\zsem$ and $\zsyn$ to capture semantic and syntactic information, respectively. Specifically, we assume that the probability of a sentence $\bm x$ in \method could be computed as:
\begin{equation*}
    \begin{split}
    &p(\bm x)= \int p(\zsem, \zsyn) p(\bm x|\zsem, \zsyn)\intd\!\zsem \intd\! \zsyn \\
    & = \int p(\zsem) p(\zsyn) p(\bm x|\zsem, \zsyn)\intd\!\zsem \intd\! \zsyn
    \end{split}
\end{equation*}
where $p(\zsem)$ and $p(\zsyn)$ are the priors; both are set to be independent multivariate Gaussian $\mathcal{N}(\bm 0, \rm I)$.

Similar to~(\ref{eqn:elbo}), we  optimize the \textit{evidence lower bound}~(ELBO) for training:
\begin{equation}
\begin{split}
    \log p(\bm x) &\ge \operatorname{ELBO}\\ \nonumber
    &= \E\limits_{q(\zsem |\bm x) q(\zsyn | \bm x)}\big[\log\, p(\bm x|\zsem,\zsyn)\big]\\ \nonumber
    &\quad- \KL\left(q(\zsem|\bm x)\; \big\| \; p(\zsem)\right)\\
    &\quad- \KL\left(q(\zsyn|\bm x)\; \big\| \; p(\zsyn)\right) 
\end{split}
\end{equation} 
where $q(\zsem|\bm x)$ and $q(\zsyn|\bm x)$ are posteriors for the two latent variables.
We further assume the variational posterior families, $q(\zsem|\bm x)$ and  $q(\zsyn|\bm x)$, are independent, taking the form $\mathcal{N}(\bm\mu_{\text{sem}}, \bm\sigma_{\text{sem}}^{2})$ and $\mathcal{N}(\bm\mu_{\text{syn}}, \bm\sigma_{\text{syn}}^{2})$, respectively, We use RNN to parameterize the posteriors (also called the \textit{encoder}).
Here, $\bm\mu_{\text{sem}}$, $\bm\sigma_{\text{sem}}$, $\bm\mu_{\text{syn}}$, and $\bm\sigma_{\text{syn}}$ are predicted by the encoder network, described as follows. 

\paragraph{Encoding}  In the encoding phase, we first obtain the sentence representation $\bm r_x$ by an RNN with the gated recurrent units~\cite[GRUs,][]{gru}; then, $\bm r_x$ is evenly split into two spaces $\bm r_x = [\bm r_x^\text{sem};\bm r_x^\text{syn}]$. 

For the semantic encoder, we compute the mean and variance of $q(\zsem|\bm x)$ from $\bm r_x^\text{sem}$ as:
\begin{equation*}
    {\left[ \begin{array}{c}
       \bm\mu_{\text{sem}}    \\
        \bm\sigma_{\text{sem}}
    \end{array}\right]}
    = 
    {\left[ \begin{array}{c}
       W_\text{sem}^{\mu}   \\
        W_\text{sem}^{\sigma}
    \end{array}\right]} 
    \operatorname{ReLU}(W_\text{sem}  \bm r_x^\text{sem}+\bm b_\text{sem})
\end{equation*}
where the activation function is the rectified linear unit~\cite[ReLU,][]{nair2010rectified}. $W_\text{sem}^{\mu}$,$W_\text{sem}^{\sigma}$,$W_\text{sem}$, and $\bm b_\text{sem}$ are the parameters of the semantic encoder.

Likewise, a syntactic encoder predicts $\bm \mu_\text{syn}$ and $\bm\sigma_\text{syn}$ for $q(\bm z_\text{syn}|\bm x)$ in the same way, with parameters $W_\text{syn}^{\mu}$,$W_\text{syn}^{\sigma}$,$W_\text{syn}$, and $\bm b_\text{syn}$.

\paragraph{Decoding in the Training Phase} We first sample from the posterior distributions by the reparameterization trick~\cite{vae}, obtaining sampled semantic and syntactic representations, ${\zsem}$ and ${\zsyn}$;  then, they are concatenated as $\bm z = [{\zsem}; \zsyn] $ and fed as the initial state of the decoder for reconstruction.

\paragraph{Decoding in the Test Phase} The treatment depends on applications. If we would like to synthesize a sentence from scratch, both $\zsyn$ and $\zsem$ are  sampled from prior. If we would like to preserve/vary semantics/syntax, max a posterior~(MAP) inference or sampling could be applied in respective spaces. Details are provided in \S~\ref{sec:exp}.


In the following part, we will introduce how syntax is modeled in our approach and how syntax and semantics are ensured to be separated.

\subsubsection{Modeling Syntax by Predicting Linearized Tree Sequence}\label{sss:syntax}

\begin{figure}
\centering
\includegraphics[width=0.40\textwidth]{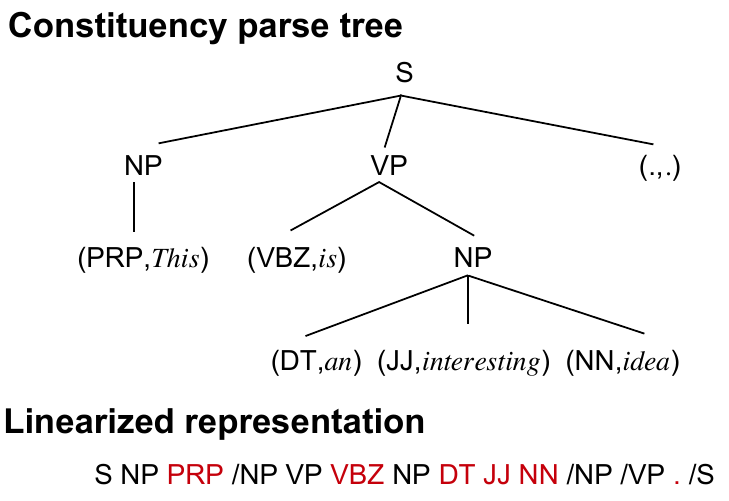}
\caption{The parse tree and its linearized tree sequence of a sentence ``\textit{This is an interesting idea.}''}
\label{fig:tree}
\end{figure}
While previous studies have tackled the problem of categorical sentiment modeling in the latent space~\cite{hu,fu}, syntax is much more complicated and  not finitely categorical. 
We propose to adopt the linearized tree sequence to explicitly model syntax in the latent space of VAE.

Figure~\ref{fig:tree} shows the constituency parse tree of the sentence ``\textit{This is an interesting idea}.'' The linearized tree sequence can be obtained by traversing the syntactic tree in a top-down order; if the node is non-terminal, we add a backtracking node~(e.g., \texttt{/NP}) after its child nodes are traversed. 

We ensure that $\zsyn$ contains syntactic information by predicting the linearized tree sequence. In training, the parse tree for sentences are obtained by the \texttt{ZPar}\footnote{\url{https://www.sutd.edu.sg/cmsresource/faculty/yuezhang/zpar.html}} toolkit, and serves as the groundtruth training signals;  in testing, we do not need external syntactic trees. We build an RNN (independent of the VAE's decoder) to predict such linearized parse trees, where each parsing token is represented by an embedding (similar to a traditional RNN decoder). 
Notice that, a node and its backtracking, e.g., \texttt{NP} and \texttt{/NP}, have different embeddings.

\begin{figure}[!t]
\centering
\includegraphics[width=0.45\textwidth]{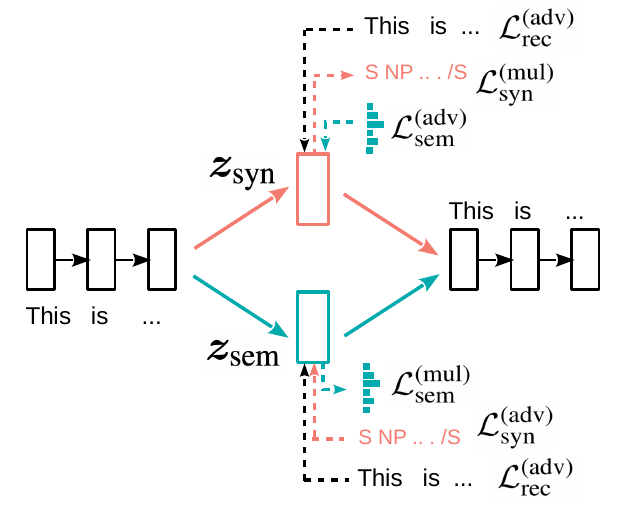}
\caption{Overview of our \methodp. Forward dashed arrows are multi-task losses; backward dashed arrows are adversarial losses.}
\label{fig:overview}
\end{figure}

The linearized tree sequence has achieved promising parsing results in a traditional constituency parsing task~\cite{vinyals2015,liu2018improving,vaswani2017attention}, which shows its ability of preserving syntactic information.
Additionally, the linearized tree sequence works in a sequence-to-sequence fashion, so that it can be used to regularize the latent spaces.

\subsubsection{Disentangling Syntax and Semantics into Different Latent Spaces}\label{sss:vineet}

Having solved the problem of syntactic modeling, we now turn to the question: how could we disentangle syntax and semantics from each other?

We are inspired by the research in text style transfer and apply auxiliary losses to regularize the latent space~\cite{hu,fu}. 

In particular, we adopt the multi-task and adversarial losses in~\newcite{vineet}, but extend it to the sequence level. In \S\ref{sss:advrec}, we further propose two adversarial reconstruction losses to discourage the model to encode a sentence from a single subspace.

\paragraph{Multi-Task Loss}Intuitively, a multi-task loss ensures that each space~($\zsyn$ or $\zsem$) should capture respective information.


For the semantic space, we predict the bag-of-words~(BoW) distribution of a sentence from $\zsem$ with softmax, whose objective is the cross-entropy loss against the groundtruth distribution $\bm t$, given by:
\begin{equation}\label{eqn:mul_sem}
    \mathcal L_\text{sem}^\text{(mul)}= - \sum\nolimits_{w\in\mathcal V}t_w \log p(w|\zsem)
\end{equation}
where $p(w|\zsyn)$ is the predicted distribution.
BoW has been explored by previous work~\cite{weng2017neural,vineet}, showing good ability of preserving semantics.

For the syntactic space, the multi-task loss trains a model to predict syntax on $\zsyn$. Due to our proposal in \S\ref{sss:syntax}, we could build a dedicated RNN, predicting the tokens in the linearized parse tree sequence, whose loss is:
\begin{equation}\label{eqn:mul_syn}
\mathcal{L}_\text{syn}^\text{(mul)}= -\sum\nolimits_{i=1}^n \log p(s_i|s_1\cdots s_{i-1}, \zsyn)
\end{equation}
where $s_i$ is a token in the linearized parse tree (with a total length of $n$). 

\paragraph{Adversarial Loss} The adversarial loss is widely used for aligning samples from different distributions. It has various applications, including style transfer~\cite{hu,fu,vineet} and domain adaptation~\cite{advDA}. 
To apply adversarial losses, we add extra model components (known as \textit{adversaries}) to predict semantic information $t_w$ based on the syntactic space $\zsyn$, but to predict syntactic information $s_1\cdots s_{n-1}$ based on the semantic space $\zsem$. They are denoted by $p_\text{adv}(w|\zsyn)$ and $p_\text{adv}(s_i|s_1\cdots s_{i-1},\zsem)$. 

The training of these adversaries are similar to (\ref{eqn:mul_sem}) and (\ref{eqn:mul_syn}), except that the gradient only trains the adversaries themselves, and does not back-propagate to VAE.

Then, VAE is trained to ``fool'' the adversaries by maximizing their losses, i.e., minimizing the following terms:
\begin{align}\label{eqn:adv_sem}
    \mathcal L_\text{sem}^\text{(adv)}&= \sum\nolimits_{w\in\mathcal V} t_w\log p_\text{adv}(w|\zsyn)\\\label{eqn:adv_syn}
    \mathcal L_\text{syn}^\text{(adv)}&= \sum\nolimits_{i=1}^n\log p_\text{adv}(s_i|s_1\cdots s_{i-1},\zsem)
\end{align}
In this phase, the adversaries are fixed and their parameters are not updated.

\subsubsection{Adversarial Reconstruction Loss}\label{sss:advrec}

Our next intuition is that syntax and semantics are more interwoven to each other than other information such as style and content. 

Suppose, for example, the syntax and semantics have been perfectly separated by the losses in \S\ref{sss:vineet}, where $\zsem$ could predict BoW well, but does not contain any information about the syntactic tree. 
Even in this ideal case, the decoder can reconstruct the original sentence from $\zsem$ by simply learning to re-order words (as $\zsem$ does contain BoW). 
Such word re-ordering knowledge is indeed learnable~\cite{ma2018bag}, and does not necessarily contain the syntactic information. 
Therefore, the multi-task and adversarial losses for syntax and semantics do not suffice to regularize DSS-VAE.

We now propose an \textit{adversarial reconstruction loss} to discourage the sentence being predicted by a single subspace $\zsyn$ or $\zsem$. 
When combined, however, they should provide a holistic view of the entire sentence. 
Formally, let $\bm z_s$ be a latent variable ($\bm z_s=\zsyn$ or $\zsem$). A decoding adversary is trained to predict the sentence based on $\bm z_s$, denoted by $p_\text{rec}(x_i|x_1\cdots x_{i-1}, \bm z_s)$. 
Then, the adversarial reconstruction loss is imposed by minimizing
\begin{align}\label{eqn:rec_loss}
    \mathcal{L}^\text{(adv)}_\text{rec}(\bm z_s)=\sum\nolimits_{i=1}^M\log p_\text{rec}(x_i|x_{<i},\bm z_s)
\end{align}
Such adversarial reconstruction loss is applied to both the syntactic and semantic spaces, shown by black bashed arrows in Figure~\ref{fig:overview}.

\subsection{Training Details}\label{ss:train}
\paragraph{Overall Training Objective}
The overall training loss is a combination of the VAE loss~(\ref{eqn:elbo}), the multi-task and adversarial losses for syntax and semantics ~(\ref{eqn:mul_sem}--\ref{eqn:adv_syn}), as well as the adversarial reconstruction losses~(\ref{eqn:rec_loss}), , i.e., minimizing 
\begin{equation}\label{eqn:overall}
\begin{split}
    \mathcal{L} &= \mathcal{L}_\text{vae} + \mathcal{L}_\text{aux} \\
    &= -\E\limits_{q(\zsem |\bm x) q(\zsyn | \bm x)}\log \big[p(\bm x|\zsem,\zsyn)\big]\\
    &\quad+ \lambda^\text{KL}_\text{sem} \KL\left(q(\zsem|\bm x)\; \big\| \; p(\zsem)\right)\\
    &\quad+ \lambda^\text{KL}_\text{syn}\KL\left(q(\zsyn|\bm x)\; \big\| \; p(\zsyn)\right) \\
    &\quad +\lambda^\text{mul}_\text{sem} \mathcal{L}^\text{(mul)}_\text{sem} +\lambda^\text{adv}_\text{sem} \mathcal{L}^\text{(adv)}_\text{sem}+\lambda^\text{rec}_\text{sem} \mathcal{L}^\text{(adv)}_\text{rec}(\zsem)\\
    &\quad +\lambda^\text{mul}_\text{syn} \mathcal{L}^\text{(mul)}_\text{syn} +\lambda^\text{adv}_\text{syn} \mathcal{L}^\text{(adv)}_\text{syn} +\lambda^\text{rec}_\text{syn} \mathcal{L}^\text{(adv)}_\text{rec}(\zsyn)\\
    \end{split}
\end{equation}
where the $\lambda^\text{KL}_\text{sem} $, $\lambda^\text{KL}_\text{syn} $, 
$\lambda^\text{mul}_\text{sem} $, $\lambda^\text{adv}_\text{sem} $, $\lambda^\text{rec}_\text{sem} $, $\lambda^\text{mul}_\text{syn} $, $\lambda^\text{adv}_\text{syn} $, and $\lambda^\text{rec}_\text{syn} $ are the hyperparameters to adjust the importance of each loss in overall objective. 

\paragraph{Hyperparameter Tuning}
We select the parameter values with the lowest ELBO value on the validation set in all experiments. 
They are tuned by~(grouped) grid search on the validation set, but due to the large hyperparameter space, we conduct tuning mostly for sensitive hyperparameters and admit that it is empirical. 
We choose the VAE as our baseline, and the KL weight of VAE is tuned in the same way. 
We list the hyperparameters in Appendix~\ref{ss:details}.
 
The training objective is optimized by Adam~\cite{adam} with $\beta_{1}$ = $0.9$, $\beta_{2}$ = $0.995$, and the initial learning rate is 0.001. Word embeddings are  300-dimensional and initialized randomly. The dimension of each latent space (namely, $\zsyn$ and $\zsem$)  is 100. 

\paragraph{KL Annealing and Word Dropout}We adopt the tricks of KL annealing and word dropout from~\newcite{bowman} to avoid KL collapse. 
We anneal $\lambda^\text{KL}_\text{syn}$ and $\lambda^\text{KL}_\text{syn}$ from zero to predefined values in a sigmoid manner.
Besides, the word dropout trick randomly replaces the ground-truth token with  $<$unk$>$ with a fixed probability of  0.50 at each time step of the decoder during training. 

\section{Experiments}\label{sec:exp}
We evaluate our method on reconstruction and unconditional language generation (\S\ref{ss:LM}). Then, we apply it two applications, namely, unsupervised paraphrase generation~(\S\ref{ss:paraphrase}) and syntax-transfer generation~(\S\ref{ss:control}). 

\subsection{Reconstruction and Unconditional Language Generation}\label{ss:LM}
First, we compare our model in reconstruction and unconditional language generation with a traditional VAE and a syntactic language model~\cite[PRPN,][]{shen2017neural}. 

\paragraph{Dataset} We followed previous work~\cite{bowman} and used a standard benchmark, the WSJ sections in the Penn Treebank~(PTB)~\cite{marcus1993building}. 
We also followed the standard split: Sections 2--21 for training, Section 24 for validation, and Section 23 for test.

\paragraph{Settings} We trained VAE and DSS-VAE, both with 100-dimensional RNN states. For the vocabulary, we chose 30k most frequent words. We trained PRPN with the default parameter in the code base.\footnote{\url{https://github.com/yikangshen/PRPN}} 

\paragraph{Evaluation} We evaluate model performance with the following metrics:
\begin{enumerate} 
	\item {Reconstruction BLEU.} The reconstruction task aims to generate the input sentence itself. In the task, both syntactic and semantic vectors are chosen as the predicted mean of the encoded distribution.
	We evaluate the reconstruction performance by the BLEU score~\cite{papineni2002bleu} with input as the reference.\footnote{We evaluate the corpus BLEU implemented in  \url{https://www.nltk.org/\_modules/nltk/translate/bleu\_score.html}} It reflects how well the model could preserve input information, and is crucial for representation learning and ``goal-oriented'' text generation. 
\begin{table}[!t]
\footnotesize
\centering
\begin{tabular}{c|cc} \hline
KL-Weight & BLEU$^\uparrow$ & Forward PPL$^\downarrow$ \\ \hline
1.3  & 7.26 & 34.01\\
1.2  & 7.41 & 35.00 \\
1.0  & 8.19 & 36.53 \\
0.7  & 8.98 & 42.44 \\
0.5  & 9.07 & 44.11 \\
0.3  & 9.26 & 48.70 \\
0.1  & 9.36 & 49.73 \\ \hline
\end{tabular}
\caption{BLEU and Forward PPL of VAE with varying KL weights on the PTB test set. The larger$^\uparrow$ (or lower$^\downarrow$), the better.}
\label{tab:KL}
\end{table}
	\item {Forward PPL}. We then perform unconditioned generation, where both syntactic and semantic vectors are sampled from prior. Forward perplexity (PPL)~\cite{zhao2017arae} is the generated sentences' perplexity score predicted by a pertained language model.\footnote{We used an LSTM language model trained on the One-Billion-Word Corpus (\url{http://www.statmt.org/lm-benchmark}).} It shows the fluency of generated sentences from VAE's prior. We computed Forward PPL based on 100K sampled sentences.
    \item {Reverse PPL.} Unconditioned generation is further evaluated by Reverse PPL~\cite{zhao2017arae}. It is obtained by first training a language model\footnote{Tied LSTM-LM with 300 dimensions and two layers, implemented in~\url{https://github.com/pytorch/examples/tree/master/word_language_model}} on 100K sampled sentences from a generation model; then, Reverse PPL is the perplexity of the PTB test sets with the trained language model. Reverse PPL evaluates the diversity and fluency of sampled sentences from a language generation model. If sampled sentences are of low diversity, the language model would be trained only on similar sentences; if the sampled sentences are of low fluency, the language model would be trained on unfluent sentences. Both will lead to higher Reverse PPL. For comparing VAE and DSS-VAE, we sample latent variables from the prior, and feed them to the decoder for generation; for LSTM-LM, we first feed the start sentence token $<$s$>$ to the decoder, and sample the word at each time step by  predicted probabilities~(i.e., forward sampling).
\end{enumerate}

\begin{figure}[!t]
\centering
\includegraphics[width=0.5\textwidth]{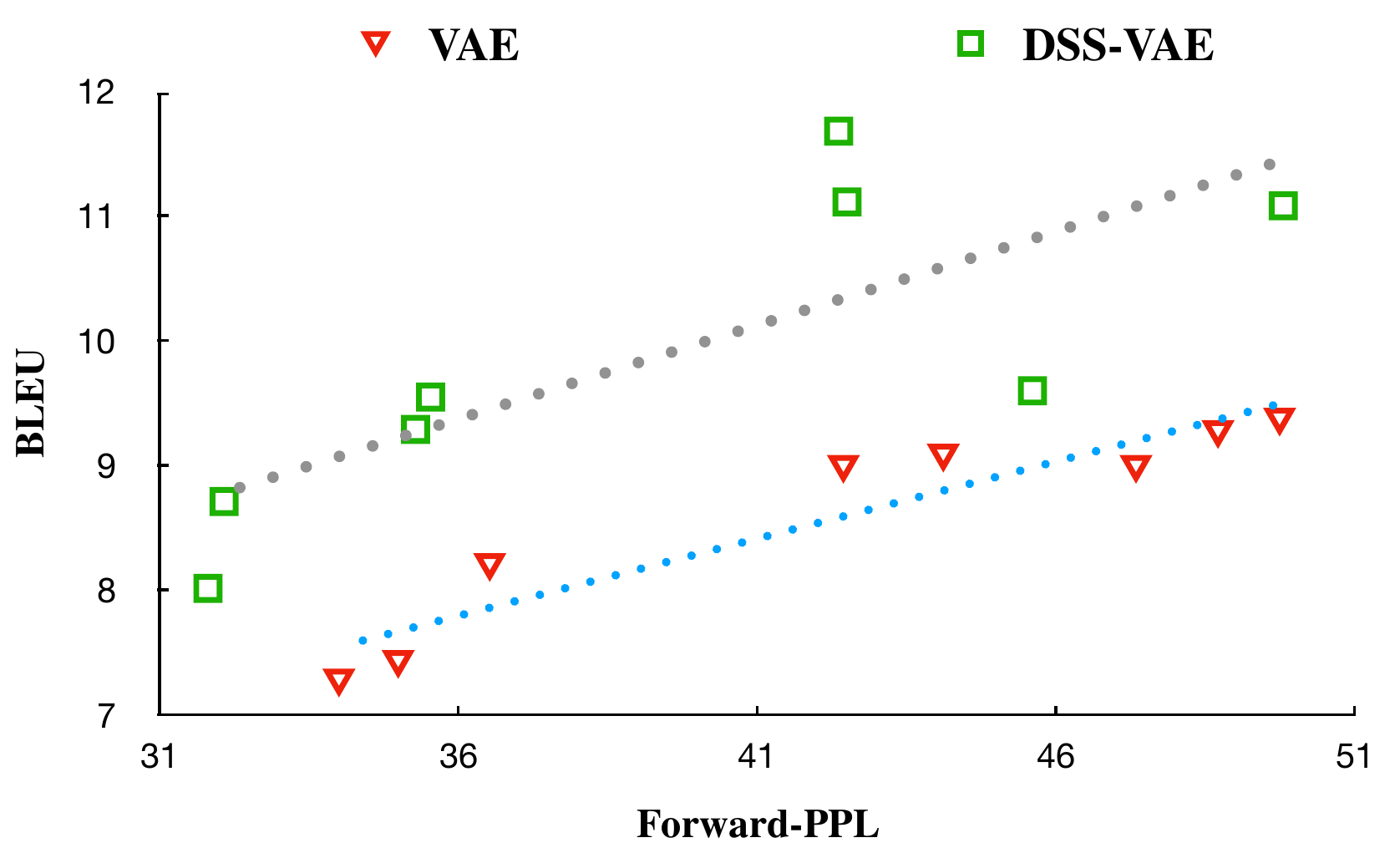}
\caption{Comparing \method and VAE in language generation with different KL weight. We performed linear regression for each model to show the trend. The upper-left corner (larger BLEU but smaller PPL) indicates a better performance.}
\label{fig:generation}
\end{figure}

\paragraph{Results} We see in Table~\ref{tab:KL} that BLEU and PPL are more or less contradictory. Usually, a smaller KL weight makes the autoencoder less ``variational'' but more ``deterministic,'' leading to less fluent sampled sentences but better reconstruction. If the trade-off is not analyzed explicitly, the VAE variant could have arbitrary results based on KL-weight tuning, which is unfair. 

We therefore present the scatter plot in Figure~\ref{fig:generation}, showing the trend of forward PPL and BLEU scores with different KL weights. 
Clearly, \method outperforms a plain VAE in BLEU if Forward PPL is controlled, and in Forward PPL if BLEU is controlled. The scatter plot shows that our proposed \method outperforms the original counterpart in language generation with different KL weights. 

In terms of Reverse PPL (Table~\ref{tab:ReversePPL}), \method also achieves better Reverse PPL than a traditional VAE.  Since \method leverages syntax to improve the sentence generation, we also include a state-of-the-art syntactic language model~\cite[PRPN-LM,][]{shen2017neural} for comparison. Results show that \method has achieved a Reverse PPL comparable to (and slightly better than) PRPN-LM. It is also seen that explicitly  modeling syntactic structures does yield better generation results---\method and PRPN consistently outperform VAE and LSTM-LM in sentence generation. 

We also include the Reverse PPL of the real training sentences. As expected, training a language model on real data outperforms training on sampled sentences from a generation model, showing that there is still much  room for improvement for all current sentence generators.

\begin{table}[!t]
\footnotesize
\centering
\begin{tabular}{l|c} \hline 
Model & Reverse PPL$^\downarrow$ \\ \hline
Real data  & 70.76 \\ 
LSTM-LM & 132.46 \\
PRPN-LM & 116.67 \\
VAE  & 125.86 \\ 
\methodp & \textbf{116.23} \\ \hline
\end{tabular}
\caption{Reverse PPL reflect the diversity and fluency of sampling data, the lower$^\downarrow$, the better. Training on the model sampled and evaluated on the real test set. We set the same KL weight for \method and VAE here.(KL weight=1.0) }
\label{tab:ReversePPL}
\end{table}

\subsection{Unsupervised Paraphrase Generation}\label{ss:paraphrase}
Given an input sentence, paraphrase generation aims to synthesize a sentence that appears different from the input, but conveys the same meaning. 
We propose a novel approach to unsupervised paraphrase generation with DSS-VAE. 
Suppose a \method is well trained according to~\S\ref{ss:train}, our approach works in the inference stage.

For a particular input sentence $\bm x^*$, let $q(\bm z_\text{syn}|\bm x^*)$ and $q(\bm z_\text{sem}|\bm x^*)$ be the encoded posterior distributions of the syntactic and semantic spaces, respectively. The inferred latent vectors are:
\begin{align}
\bm z^*_\text{sem}&=\operatorname*{argmax}\nolimits_{\bm z_\text{sem}}\  q(\bm z_\text{sem}|\bm x^*)\\
\bm z^*_\text{syn}&\sim  q(\bm z_\text{syn}|\bm x^*)     
\end{align}
and are further combined as:
\begin{equation}\label{eqn:combine}
    \bm z^* = \left[\bm z^*_\text{syn}; \bm z^*_\text{sem}\right]
\end{equation}
Finally, $\bm z^*$ is fed to the decoder and perform a greedy decoding for paraphrase generation.

The intuition behind is that, when generating the paraphrase, semantics should remain the same, but the syntax of a paraphrase could (and should) vary. 
Therefore, we sample a $\bm z^*_\text{syn}$ vector from its probabilistic distribution, while fixing $\bm z^*_\text{sem}$. 

\paragraph{Dataset}We used the established Quora dataset\footnote{\url{https://www.kaggle.com/c/quora-question-pairs/data}} to evaluate paraphrase generation, following previous work~\cite{MH}. The dataset contains 140k pairs of paraphrase sentences and 260k pairs of non-paraphrase sentences. In the standard dataset split, there are 3k and 30k held-out validation and test sets, respectively. In this experiment, we consider the unsupervised setting as~\newcite{MH}, using all non-paraphrase sentences as training samples. It is also noted that we only valid our model on the non-paraphrase held-out validation set by selecting with  the lowest validation ELBO.

\paragraph{Evaluation}
Since the test set contains a reference paraphrase for each input, it is straightforward to compute the BLEU against the reference, denoted by BLEU-ref. However, this metric alone does not model whether the generated sentence is different from the input, and thus, \newcite{MH} propose to measure this by computing BLEU against the original sentence (denoted as BLEU-ori), which ideally should be low. We only consider the \method that yields a BLEU-ori lower than 55, which is empirically suggested by~\newcite{MH} that ensures the obtained sentence is different from the original to at least a certain degree. 

\begin{table}[!t]
\centering
\footnotesize
\begin{tabular}{l|cc} \hline
Model & BLEU-ref$^\uparrow$\!\! & BLEU-ori$^\downarrow$ \\ \hline
Origin Sentence$^\dag$ & 30.49 & 100 \\
VAE-SVG-eq (supervised)$^\ddag$	& 22.90& -- \\ 
VAE (unsupervised)$^\dag$& 9.25 & 27.23 \\ 
CGMH$^\dag$ & 18.85 & 50.18 \\ 
\methodp & \textbf{20.54} & 52.77 \\ \hline
\end{tabular}
\caption{Performance of paraphrase generation. The larger$^\uparrow$ (or lower$^\downarrow$), the better. Some results are quoted from $^\dag$\newcite{MH} and $^\ddag$\newcite{vae_svg_eq}.}
\label{tab:paraphrase}
\end{table}

\paragraph{Results} Table~\ref{tab:paraphrase} shows the performance of unsupervised paraphrase generation. In the first row of Table~\ref{tab:paraphrase}, simply copying the original sentences yields the highest BLEU-ref, but is meaningless as it has a BLEU-ori score of 100. We see that \method outperforms the CGMH and the original VAE in BLEU-ref. Especially, \method achieves a closer BLEU-ref compared with supervised paraphrase methods~\cite{vae_svg_eq}.

\begin{figure}[!tb]
\centering
\includegraphics[width=0.5\textwidth]{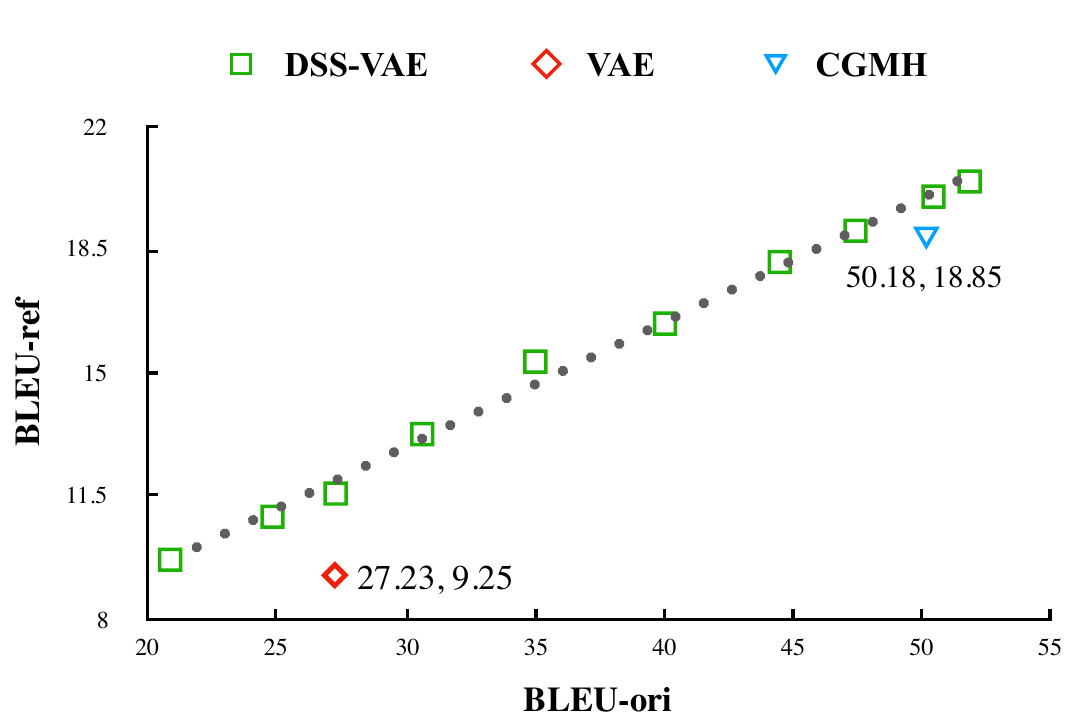}
\caption{Trade-off between BLEU-ori (the lower, the better) and BLEU-ref (the larger, the better) in unsupervised paraphrase generation. Again, the upper-left corner indicates a better performance.}
\label{fig:up}
\end{figure}

We admit that it is hard to present the trade-off by listing a single score for each model in the Table~\ref{tab:paraphrase}. We therefore have the scatter plot in Figure~\ref{fig:up} to further compare these methods. As seen, the trade-off is pretty linear and less noisy compared with Figure~\ref{fig:generation}. It is seen that the line of \method is located to the upper-left of the competing methods. 
In other words, the plain VAE and CGMH are ``inadmissible,'' meaning that \method simultaneously outperforms them in both BLEU-ori and BLEU-ref, indicating that \method outperforms previous state-of-the-art methods in unsupervised paraphrase generation.

\subsection{Syntax-Transfer Generation}\label{ss:control} 
In this experiment, we propose a novel application of syntax-transfer text generation, inspired by previous sentiment-style transfer studies~\cite{hu,fu,vineet}.
\begin{table*}[!t]
\centering
\resizebox{\linewidth}{!}{%
\begin{tabular}{l|ccc|ccc|c} \hline
\multicolumn{1}{c|}{\multirow{2}{*}{Model}} & \multicolumn{2}{c}{word-BLEU (corpus)}& \multirow{2}{*}{$\Delta \text{word-BLEU}{}^\uparrow$}& \multicolumn{2}{|c}{Average TED (per sentence)}& \multirow{2}{*}{$\Delta$TED${}^\uparrow$} & \multirow{2}{*}{Geo Mean $\Delta{}^\uparrow$}\\ 
\multicolumn{1}{c|}{} & $\refsem{}^\uparrow$& $\refsyn{}^\downarrow$ &  & $\refsem{}^\uparrow$& $\refsyn{}^\downarrow$ & &  \\ \hline 
VAE  & 6.81 & 6.68 & 0.13 & 149.22 & 148.59 & 0.63 & 0.29 \\ \hline
$\mathcal L_\text{sem}^\text{(mul)}+\mathcal L_\text{syn}^\text{(mul)}+\mathcal L_\text{sem}^\text{(adv)}+\mathcal L_\text{syn}^\text{(adv)}$
& 12.14 & 6.22 & 5.92 & 159.51 & 134.80 & 24.71 & 12.09 \\
\quad $+ \mathcal{L}^\text{(adv)}_\text{rec}(\bm \zsem)$
& 11.83 & 6.60 & 5.23 & \textbf{163.40} & \textbf{131.27} & \textbf{32.13} & 12.96\\
\quad $+ \mathcal{L}^\text{(adv)}_\text{rec}(\bm \zsyn)$ 
&\textbf{14.33} & \textbf{6.07} & \textbf{8.26} & 159.20 & 134.22 & 24.98 & 14.36\\
\quad  $+ \mathcal{L}^\text{(adv)}_\text{rec}(\bm \zsyn)+\mathcal{L}^\text{(adv)}_\text{rec}(\bm \zsem)$ 
& 13.74 & 6.15 & 7.59 & 161.94 & 131.09 & 30.85 & \textbf{15.30} \\ \hline
\end{tabular}%
}
\caption{Performance of syntax-transfer generation. The larger$^\uparrow$ (or lower$^\downarrow$), the better. The results of VAE are obtained by averaging interpolation. $\Delta \text{word-BLEU} = \text{word-BLEU}(\refsem) - \text{word-BLEU}(\refsyn)$. We also compute the difference as $\Delta \text{TED} = \text{TED}(\refsem) - \text{TED}(\refsyn)$ to measure if the generated sentence is syntactically similar to $\refsyn$ but not $\refsem$. Due to the difference of scale between BLEU and TED, we compute the geometric mean of $\Delta \text{word-BLEU}$ and $\Delta \text{TED}$ reflect the total differences.}
\label{tab:transfer}
\end{table*}

Consider two sentences:\\[-0.6cm]
\begin{enumerate}
\item[$\bm x_1$:] \texttt{\small There is a dog behind the door.} \\[-0.6cm]
\item[$\bm x_2$:] \texttt{\small The child is playing in the garden.}
\end{enumerate}
If we would like to generate a sentence having the syntax of ``\texttt{there is/are}'' as $\bm x_1$ but conveying the meaning of $\bm x_2$, we could graft the respective syntactic and semantic vectors as:
\begin{align*}
     \bm z^*_\text{sem} &= \operatorname{argmax}_{\bm z_\text{sem}}q(\bm z_\text{sem}|\bm x_2) \\
    \bm z^*_\text{syn} &= \operatorname{argmax}_{\bm z_\text{syn}}\ \ \ q(\bm z_\text{syn}|\bm x_1) \\
    \bm z &= \left[\bm z^*_\text{sem}; \bm z^*_\text{syn}\right]
\end{align*}
and then feed $\bm z$ to the decoder to obtain a syntax-transferred sentence. 

\paragraph{Dataset and Evaluation} To evaluate this task, we  constructed a  subset of the Stanford Natural Language Inference (SNLI), containing 1000 non-paraphrase pairs.  SNLI sentences can be thought of as a simple domain-specific corpus, but were all written by humans.  
In each pair we constructed, one sentence serves as the semantic provider (denoted by $\refsem$), and the other serves as the syntactic provider (denoted by $\refsyn$). The goal of syntax-transfer text generation is to synthesize a sentence that resembles $\refsem$ but not $\refsyn$ in semantics, and resembles $\refsyn$ but not $\refsem$ in syntax. For the semantic part, we use the traditional word-based BLEU scores to evaluate how the generated sentence is close to  $\refsem$  but different from $\refsyn$.
For syntactic similarity, we use the zss package\footnote{\url{https://github.com/timtadh/zhang-shasha}} to calculate the Tree Edit Distance~\cite[TED,][]{zhang1989simple}. TED is essentially the minimum-cost sequence of node edit operations (namely, delete, insert, and rename) between two trees, which reflects the difference of two syntactic trees. 

Since we hope the generated sentence has a higher word-BLEU score compared with Ref$_\text{sem}$ but a lower word-BLEU score compared with Ref$_\text{syn}$, we compute their difference, denoted by $\Delta\text{word-BLEU}$, to consider both. Likewise, $\Delta$TED is also computed. We further take the geometric mean of $\Delta\text{word-BLEU}$ and $\Delta\text{TED}$ to take both into account.

\paragraph{Results} We see from Table~\ref{tab:transfer} that a traditional VAE cannot accomplish the task of syntax transfer. This is because $\refsyn$ and $\refsem$---even if we artificially split the latent space into two parts---play the same role in the decoder. 
With the multi-task and adversarial losses for syntactic and semantic latent spaces, the total difference is increased by 12.09, which shows the success of syntax-transfer sentence generation.
This further implies that explicitly modeling syntax is feasible in the latent space of VAE.
We incrementally applied the adversarial reconstruction loss, proposed in \S~\ref{sss:advrec}.

As seen, an adversarial reconstruction loss drastically strengthens the role of the other space. For example, $+ \mathcal{L}^\text{(adv)}_\text{rec}(\bm \zsem)$  repels information to  the syntactic space and  achieves the highest $\Delta$TED.

When applying the adversarial reconstruction losses to both semantic and syntactic spaces, we have a balance between $\Delta$word-BLEU and $\Delta$TED, both ranking second in the respective columns. Eventually, we achieve the highest total difference, showing that our full \method model achieves the best performance of syntax-transfer generation.

\paragraph{Discussion on syntax transfer between incompatible sentences}
We provide a few case studies of syntax-transfer generation in Appendix~\ref{sec:case}. 
We empirically find that the syntactic transfer between ``compatible'' sentences give more promising results than transfer between ``incompatible'' sentences.
Intuitively, this is  reasonable because it may be  hard to transfer a sentence with a length of 5, say, to a sentence with a length of 50.

\section{Conclusion}
In this paper, we propose a novel DSS-VAE model, which explicitly models syntax in the distributed latent space of VAE and  enjoys the benefits of sampling and manipulation in terms of the syntax of a sentence. 
Experiments show that \method outperforms the VAE baseline in reconstruction and unconditioned language generation.
We further make use of the sampling and manipulation advantages of \method in two novel applications, namely unsupervised paraphrase and syntax-transfer generation. In both experiments,  \method achieves promising results.

\section*{Acknowledgments}
We would like to thank the anonymous reviewers for their insightful comments. 
This work is supported by the National Science Foundation of China (No.~61772261 and No.~61672277) and the Jiangsu Provincial Research Foundation for Basic Research (No.~BK20170074). 

\bibliography{acl2019}

\begin{thebibliography}{39}
\expandafter\ifx\csname natexlab\endcsname\relax\def\natexlab#1{#1}\fi

\bibitem[{Bowman et~al.(2016)Bowman, Vilnis, Vinyals, Dai, Jozefowicz, and
  Bengio}]{bowman}
Samuel~R. Bowman, Luke Vilnis, Oriol Vinyals, Andrew Dai, Rafal Jozefowicz, and
  Samy Bengio. 2016.
\newblock \href {https://doi.org/10.18653/v1/K16-1002} {Generating sentences
  from a continuous space}.
\newblock In \emph{CoNLL}, pages 10--21.

\bibitem[{Charniak(2001)}]{charniak2001immediate}
Eugene Charniak. 2001.
\newblock \href {https://doi.org/10.3115/1073012.1073029} {Immediate-head
  parsing for language models}.
\newblock In \emph{ACL}, pages 124--131.

\bibitem[{Chelba(1997)}]{chelba1997}
Ciprian Chelba. 1997.
\newblock \href {https://doi.org/10.3115/976909.979681} {A structured language
  model}.
\newblock In \emph{ACL}, pages 498--500.

\bibitem[{Chen et~al.(2017)Chen, Huang, Chiang, and Chen}]{chen2017improved}
Huadong Chen, Shujian Huang, David Chiang, and Jiajun Chen. 2017.
\newblock \href {http://www.aclweb.org/anthology/P17-1177} {Improved neural
  machine translation with a syntax-aware encoder and decoder}.
\newblock In \emph{ACL}, pages 1936--1945.

\bibitem[{Cho et~al.(2014)Cho, van Merrienboer, Gulcehre, Bahdanau, Bougares,
  Schwenk, and Bengio}]{gru}
Kyunghyun Cho, Bart van Merrienboer, Caglar Gulcehre, Dzmitry Bahdanau, Fethi
  Bougares, Holger Schwenk, and Yoshua Bengio. 2014.
\newblock \href {http://www.aclweb.org/anthology/D14-1179} {Learning phrase
  representations using {RNN} encoder--decoder for statistical machine
  translation}.
\newblock In \emph{EMNLP}, pages 1724--1734.

\bibitem[{Clark(2001)}]{clark2001unsupervised}
Alexander Clark. 2001.
\newblock \href {https://www.aclweb.org/anthology/W01-0713} {Unsupervised
  induction of stochastic context-free grammars using distributional
  clustering}.
\newblock In \emph{Proceedings of the {ACL} 2001 Workshop on Computational
  Natural Language Learning}.

\bibitem[{Dai et~al.(2018)Dai, Tian, Dai, Skiena, and
  Song}]{syntax_directed_vae}
Hanjun Dai, Yingtao Tian, Bo~Dai, Steven Skiena, and Le~Song. 2018.
\newblock \href {https://arxiv.org/pdf/1802.08786.pdf} {Syntax-directed
  variational autoencoder for structured data}.
\newblock In \emph{ICLR}.

\bibitem[{Deriu and Cieliebak(2018)}]{syntactic_mani}
Jan~Milan Deriu and Mark Cieliebak. 2018.
\newblock \href {http://www.aclweb.org/anthology/W18-65#page=42} {Syntactic
  manipulation for generating more diverse and interesting texts}.
\newblock In \emph{Proceedings of the 11th International Conference on Natural
  Language Generation}, pages 22--34.

\bibitem[{Dyer et~al.(2016)Dyer, Kuncoro, Ballesteros, and Smith}]{rnng}
Chris Dyer, Adhiguna Kuncoro, Miguel Ballesteros, and Noah~A Smith. 2016.
\newblock \href {https://arxiv.org/abs/1602.07776} {Recurrent neural network
  grammars}.
\newblock In \emph{NAACL}, pages 199--209.

\bibitem[{Eriguchi et~al.(2016)Eriguchi, Hashimoto, and
  Tsuruoka}]{eriguchi2016tree}
Akiko Eriguchi, Kazuma Hashimoto, and Yoshimasa Tsuruoka. 2016.
\newblock \href {http://www.aclweb.org/anthology/P16-1078} {Tree-to-sequence
  attentional neural machine translation}.
\newblock In \emph{ACL}, pages 823--833.

\bibitem[{Fu et~al.(2018)Fu, Tan, Peng, Zhao, and Yan}]{fu}
Zhenxin Fu, Xiaoye Tan, Nanyun Peng, Dongyan Zhao, and Rui Yan. 2018.
\newblock \href {http://arxiv.org/abs/1711.06861} {Style transfer in text:
  Exploration and evaluation}.
\newblock In \emph{AAAI}, pages 663--670.

\bibitem[{G{\'o}mez-Bombarelli et~al.(2018)G{\'o}mez-Bombarelli, Wei, Duvenaud,
  Hern{\'a}ndez-Lobato, S{\'a}nchez-Lengeling, Sheberla, Aguilera-Iparraguirre,
  Hirzel, Adams, and Aspuru-Guzik}]{cvae}
Rafael G{\'o}mez-Bombarelli, Jennifer~N Wei, David Duvenaud, Jos{\'e}~Miguel
  Hern{\'a}ndez-Lobato, Benjam{\'\i}n S{\'a}nchez-Lengeling, Dennis Sheberla,
  Jorge Aguilera-Iparraguirre, Timothy~D Hirzel, Ryan~P Adams, and Al{\'a}n
  Aspuru-Guzik. 2018.
\newblock \href
  {https://dash.harvard.edu/bitstream/handle/1/35164975/235.pdf?sequence=1&isAllowed=y}
  {Automatic chemical design using a data-driven continuous representation of
  molecules}.
\newblock \emph{ACS Central Science}, 4(2):268--276.

\bibitem[{Gupta et~al.(2018)Gupta, Agarwal, Singh, and Rai}]{vae_svg_eq}
Ankush Gupta, Arvind Agarwal, Prawaan Singh, and Piyush Rai. 2018.
\newblock \href {https://arxiv.org/pdf/1709.05074.pdf} {A deep generative
  framework for paraphrase generation}.
\newblock In \emph{AAAI}, pages 5149--5156.

\bibitem[{Hu et~al.(2017)Hu, Yang, Liang, Salakhutdinov, and Xing}]{hu}
Zhiting Hu, Zichao Yang, Xiaodan Liang, Ruslan Salakhutdinov, and Eric~P. Xing.
  2017.
\newblock \href {http://proceedings.mlr.press/v70/hu17e.html} {Toward
  controlled generation of text}.
\newblock In \emph{ICML}, pages 1587--1596.

\bibitem[{John et~al.(2018)John, Mou, Bahuleyan, and Vechtomova}]{vineet}
Vineet John, Lili Mou, Hareesh Bahuleyan, and Olga Vechtomova. 2018.
\newblock \href {https://arxiv.org/pdf/1808.04339.pdf} {Disentangled
  representation learning for text style transfer}.
\newblock \emph{arXiv preprint arXiv:1808.04339}.

\bibitem[{Kingma and Ba(2015)}]{adam}
Diederik~P. Kingma and Jimmy Ba. 2015.
\newblock \href {http://arxiv.org/abs/1412.6980} {Adam: A method for stochastic
  optimization}.
\newblock In \emph{ICLR}.

\bibitem[{Kingma and Welling(2014)}]{vae}
Diederik~P Kingma and Max Welling. 2014.
\newblock \href {https://arxiv.org/pdf/1312.6114.pdf} {Auto-encoding
  variational {Bayes}}.
\newblock In \emph{ICLR}.

\bibitem[{Kuncoro et~al.(2017)Kuncoro, Ballesteros, Kong, Dyer, Neubig, and
  Smith}]{whatrnng}
Adhiguna Kuncoro, Miguel Ballesteros, Lingpeng Kong, Chris Dyer, Graham Neubig,
  and Noah~A Smith. 2017.
\newblock \href {https://aclweb.org/anthology/E17-1117} {What do recurrent
  neural network grammars learn about syntax?}
\newblock In \emph{EACL}, pages 1249--1258.

\bibitem[{Kusner et~al.(2017)Kusner, Paige, and Hern{\'a}ndez-Lobato}]{gvae}
Matt~J. Kusner, Brooks Paige, and Jos{\'e}~Miguel Hern{\'a}ndez-Lobato. 2017.
\newblock \href {http://proceedings.mlr.press/v70/kusner17a/kusner17a.pdf}
  {Grammar variational autoencoder}.
\newblock In \emph{ICML}, pages 1945--1954.

\bibitem[{Li et~al.(2018{\natexlab{a}})Li, Jia, He, and Liang}]{percy}
Juncen Li, Robin Jia, He~He, and Percy Liang. 2018{\natexlab{a}}.
\newblock \href {http://www.aclweb.org/anthology/N18-1169} {{Delete, Retrieve,
  Generate}: a simple approach to sentiment and style transfer}.
\newblock In \emph{ACL}, pages 1865--1874.

\bibitem[{Li et~al.(2017)Li, Xiong, Tu, Zhu, Zhang, and Zhou}]{li2017modeling}
Junhui Li, Deyi Xiong, Zhaopeng Tu, Muhua Zhu, Min Zhang, and Guodong Zhou.
  2017.
\newblock \href {http://www.aclweb.org/anthology/P17-1064} {Modeling source
  syntax for neural machine translation}.
\newblock In \emph{ACL}, pages 688--697.

\bibitem[{Li et~al.(2018{\natexlab{b}})Li, Song, Zhang, Chen, Shi, Zhao, and
  Yan}]{li2018generating}
Juntao Li, Yan Song, Haisong Zhang, Dongmin Chen, Shuming Shi, Dongyan Zhao,
  and Rui Yan. 2018{\natexlab{b}}.
\newblock \href {http://www.aclweb.org/anthology/D18-1423} {Generating
  classical chinese poems via conditional variational autoencoder and
  adversarial training}.
\newblock In \emph{EMNLP}, pages 3890--3900.

\bibitem[{Liu et~al.(2018)Liu, Zhu, and Shi}]{liu2018improving}
Lemao Liu, Muhua Zhu, and Shuming Shi. 2018.
\newblock \href
  {https://www.aaai.org/ocs/index.php/AAAI/AAAI18/paper/view/16347/16018}
  {Improving sequence-to-sequence constituency parsing}.
\newblock In \emph{AAAI}, pages 4873--4880.

\bibitem[{Ma et~al.(2018)Ma, Sun, Wang, and Lin}]{ma2018bag}
Shuming Ma, Xu~Sun, Yizhong Wang, and Junyang Lin. 2018.
\newblock \href {https://arxiv.org/pdf/1805.04871.pdf} {Bag-of-words as target
  for neural machine translation}.
\newblock In \emph{ACL}, pages 332--338.

\bibitem[{Marcus et~al.(1993)Marcus, Marcinkiewicz, and
  Santorini}]{marcus1993building}
Mitchell~P Marcus, Mary~Ann Marcinkiewicz, and Beatrice Santorini. 1993.
\newblock \href {http://anthology.aclweb.org/J/J93/J93-2004.pdf} {Building a
  large annotated corpus of english: The {Penn Treebank}}.
\newblock \emph{Computational linguistics}, 19(2):313--330.

\bibitem[{Miao et~al.(2019)Miao, Zhou, Mou, Yan, and Li}]{MH}
Ning Miao, Hao Zhou, Lili Mou, Rui Yan, and Lei Li. 2019.
\newblock \href {https://arxiv.org/pdf/1811.10996.pdf} {{CGMH:} {C}onstrained
  sentence generation by {Metropolis-Hastings} sampling}.
\newblock In \emph{AAAI}.

\bibitem[{Mikolov et~al.(2010)Mikolov, Karafi{\'a}t, Burget,
  {\v{C}}ernock{\`y}, and Khudanpur}]{mikolov2010recurrent}
Tom{\'a}{\v{s}} Mikolov, Martin Karafi{\'a}t, Luk{\'a}{\v{s}} Burget, Jan
  {\v{C}}ernock{\`y}, and Sanjeev Khudanpur. 2010.
\newblock \href
  {https://www.isca-speech.org/archive/archive_papers/interspeech_2010/i10_1045.pdf}
  {Recurrent neural network based language model}.
\newblock In \emph{INTERSPEECH}.

\bibitem[{Nair and Hinton(2010)}]{nair2010rectified}
Vinod Nair and Geoffrey~E Hinton. 2010.
\newblock \href {https://www.cs.toronto.edu/~hinton/absps/reluICML.pdf}
  {Rectified linear units improve restricted boltzmann machines}.
\newblock In \emph{ICML}, pages 807--814.

\bibitem[{Papineni et~al.(2002)Papineni, Roukos, Ward, and
  Zhu}]{papineni2002bleu}
Kishore Papineni, Salim Roukos, Todd Ward, and Wei-Jing Zhu. 2002.
\newblock \href {https://www.aclweb.org/anthology/P02-1040} {{BLEU}: A method
  for automatic evaluation of machine translation}.
\newblock In \emph{ACL}, pages 311--318.

\bibitem[{Semeniuta et~al.(2017)Semeniuta, Severyn, and Barth}]{conv_vae}
Stanislau Semeniuta, Aliaksei Severyn, and Erhardt Barth. 2017.
\newblock \href {http://www.aclweb.org/anthology/D17-1066} {A hybrid
  convolutional variational autoencoder for text generation}.
\newblock In \emph{EMNLP}, pages 627--637.

\bibitem[{Serban et~al.(2017)Serban, Sordoni, Lowe, Charlin, Pineau, Courville,
  and Bengio}]{serban2017hierarchical}
Iulian~Vlad Serban, Alessandro Sordoni, Ryan Lowe, Laurent Charlin, Joelle
  Pineau, Aaron~C Courville, and Yoshua Bengio. 2017.
\newblock \href {http://www.cs.toronto.edu/~lcharlin/papers/vhred_aaai17.pdf}
  {A hierarchical latent variable encoder-decoder model for generating
  dialogues.}
\newblock In \emph{AAAI}, pages 3295--3301.

\bibitem[{Shen et~al.(2017)Shen, Lin, Huang, and Courville}]{shen2017neural}
Yikang Shen, Zhouhan Lin, Chin-Wei Huang, and Aaron Courville. 2017.
\newblock \href {http://arxiv.org/abs/1711.02013} {Neural language modeling by
  jointly learning syntax and lexicon}.
\newblock In \emph{ICLR}.

\bibitem[{Tzeng et~al.(2017)Tzeng, Hoffman, Saenko, and Darrell}]{advDA}
Eric Tzeng, Judy Hoffman, Kate Saenko, and Trevor Darrell. 2017.
\newblock \href {https://arxiv.org/pdf/1702.05464.pdf} {Adversarial
  discriminative domain adaptation}.
\newblock In \emph{CVPR}, pages 7167--7176.

\bibitem[{Vaswani et~al.(2017)Vaswani, Shazeer, Parmar, Uszkoreit, Jones,
  Gomez, Kaiser, and Polosukhin}]{vaswani2017attention}
Ashish Vaswani, Noam Shazeer, Niki Parmar, Jakob Uszkoreit, Llion Jones,
  Aidan~N Gomez, {\L}ukasz Kaiser, and Illia Polosukhin. 2017.
\newblock \href {https://arxiv.org/pdf/1706.03762v1.pdf} {Attention is all you
  need}.
\newblock In \emph{NIPS}, pages 5998--6008.

\bibitem[{Vinyals et~al.(2015)Vinyals, Kaiser, Koo, Petrov, Sutskever, and
  Hinton}]{vinyals2015}
Oriol Vinyals, {\L}ukasz Kaiser, Terry Koo, Slav Petrov, Ilya Sutskever, and
  Geoffrey Hinton. 2015.
\newblock \href
  {http://papers.nips.cc/paper/5635-grammar-as-a-foreign-language.pdf} {Grammar
  as a foreign language}.
\newblock In \emph{NIPS}, pages 2773--2781.

\bibitem[{Weng et~al.(2017)Weng, Huang, Zheng, DAI, and
  Jiajun}]{weng2017neural}
Rongxiang Weng, Shujian Huang, Zaixiang Zheng, XIN-YU DAI, and CHEN Jiajun.
  2017.
\newblock \href {https://www.aclweb.org/anthology/D17-1013} {Neural machine
  translation with word predictions}.
\newblock In \emph{EMNLP}, pages 136--145.

\bibitem[{Zhang and Shasha(1989)}]{zhang1989simple}
Kaizhong Zhang and Dennis Shasha. 1989.
\newblock \href
  {http://www.grantjenks.com/wiki/_media/ideas/simple_fast_algorithms_for_the_editing_distance_between_tree_and_related_problems.pdf}
  {Simple fast algorithms for the editing distance between trees and related
  problems}.
\newblock \emph{SIAM journal on computing}, 18(6):1245--1262.

\bibitem[{Zhao et~al.(2018)Zhao, Kim, Zhang, Rush, and LeCun}]{zhao2017arae}
Junbo~Jake Zhao, Yoon Kim, Kelly Zhang, Alexander~M. Rush, and Yann LeCun.
  2018.
\newblock \href {https://arxiv.org/pdf/1706.04223.pdf} {Adversarially
  regularized autoencoders}.
\newblock In \emph{ICML}, pages 5897--5906.

\bibitem[{Zhou et~al.(2017)Zhou, Tu, Huang, Liu, Li, and Chen}]{zhou2017chunk}
Hao Zhou, Zhaopeng Tu, Shujian Huang, Xiaohua Liu, Hang Li, and Jiajun Chen.
  2017.
\newblock \href {http://www.aclweb.org/anthology/P17-2092} {Chunk-based
  bi-scale decoder for neural machine translation}.
\newblock In \emph{ACL}, pages 580--586.

\end{thebibliography}
\bibliographystyle{acl_natbib}
\appendix
\section{Hyperparameter Details}\label{ss:details}

We list the hyperparaemters in Tables~\ref{tab:ptb} and~\ref{tab:quora}.
Every 500 batch, we save the model if it achieves a lower \textit{evidence lower bound}~(ELBO) on the validation set.
\begin{table*}[!hbt]
 \begin{minipage}{0.45\textwidth}
  \centering
  \begin{tabular}{c|c} \hline
    Hyper-parameters & Value \\ \hline
      $\lambda^\text{KL}_{sem}  $ & 1.0 \\
      $\lambda^\text{KL}_{syn}$   & 1.0 \\
      $\lambda^\text{mul}_{sem}$   & 0.5 \\
      $\lambda^\text{mul}_{syn}$   & 0.5 \\
      $\lambda^\text{adv}_{sem}$  & 0.5 \\
      $\lambda^\text{adv}_{syn}$  & 0.5 \\
      $\lambda^\text{rec}_{sem}$   & 0.5 \\
      $\lambda^\text{rec}_{syn}$   & 0.5 \\ 
      Batch size & 32 \\
      GRU Dropout  & 0.1 \\\hline
    \end{tabular}
    \caption{The hyper-parameters we used in PTB dataset}
    \label{tab:ptb}
  \end{minipage}
  \qquad 
  \begin{minipage}{0.45\textwidth}
  \centering
    \begin{tabular}{c|c} \hline
    Hyper-parameters & Value \\ \hline
      $\lambda^\text{KL}_{sem}  $ & 1/3 \\
      $\lambda^\text{KL}_{syn}$   & 2/3 \\
      $\lambda^\text{mul}_{sem}$   & 5.0 \\
      $\lambda^\text{mul}_{syn}$   & 1.0 \\
      $\lambda^\text{adv}_{sem}$  & 0.5 \\
      $\lambda^\text{adv}_{syn}$  & 0.5 \\
      $\lambda^\text{rec}_{sem}$   & 1.0 \\
      $\lambda^\text{rec}_{syn}$   & 0.05 \\ 
      Batch size & 50 \\
      GRU Dropout  & 0.3 \\ \hline
    \end{tabular}
    \caption{The hyper-parameters we used in Quora dataset.}
    \label{tab:quora}
  \end{minipage}
\end{table*}

\begin{table*}[!thb]
    \footnotesize
    \centering
    \resizebox{\textwidth}{!}{
    \begin{tabular}{l|rl} \hline 
    \multicolumn{1}{c|}
    {Semantic and Syntactic Providers} & \multicolumn{1}{c}{Syntax-Transfer Output} \\ \hline \hline
    \begin{tabular}[c]{rl}
    $\textbf{Ref}_\textbf{syn}$: & There is an apple on the table. \\
    $\textbf{Ref}_\textbf{sem}$: & The airplane is in the sky.  
    \end{tabular} & 
    \begin{tabular}[c]{rl}
    \textbf{VAE}:& The man is in the kitchen.\\ 
    \textbf{\methodp}: & \parbox{6cm}{There is a airplane in the sky.}
    \end{tabular} \\ \hline
    \begin{tabular}[c]{rl}
     $\textbf{Ref}_\textbf{syn}$: & The shellfish was cooked in a wok. \\
     $\textbf{Ref}_\textbf{sem}$: & The stadium was packed with people.
    \end{tabular} & 
    \begin{tabular}[c]{rl}
    \textbf{VAE}:& \parbox{5cm}{The man was filled with people.}\\ 
    \textbf{\methodp}:& \parbox{6cm}{The stadium was packed with people.}
    \end{tabular} \\ \hline
    \begin{tabular}[c]{rl}
    $\textbf{Ref}_\textbf{syn}$:& The child is playing in the garden. \\
    $\textbf{Ref}_\textbf{sem}$:& There is a dog behind the door.
    \end{tabular} & 
    \begin{tabular}[c]{rl}
    \textbf{VAE}:& There is a person in the garden.\\ 
    \textbf{\methodp}:& \parbox{6cm}{A dog is walking behind the door.}
    \end{tabular} \\ \hline
    \end{tabular}%
    }
    \caption{Case studies of syntax transfer generation.}
\label{tab:case}
\end{table*}

\section{Case Study of Syntax Transfer}\label{sec:case}
We  provide a few examples in Table~\ref{tab:case}. We see in all cases that a plain VAE ``interpolates'' two sentences without the consideration of syntax and semantics, whereas our \method is able to transfer the syntax without changing the meaning much. In the first example, \method successfully transfer a ``subject-be-predicative'' sentence to a ``there is/are'' sentence. 
For the second example, the semantic reference has the same syntactic structure as the syntax reference, and as a result,  \method generates the same sentence as $\refsem$. For the last example, we transfer a ``there is/are`` sentence to a ``subject-be-predicative`` sentence, and our \method is also able to generate the desired syntax.

\end{document}


\section{Appendix I: Hyper-Parameter Details}\label{ss:details}

We list the hyperparaemters in Table~\ref{tab:ptb} and Table~\ref{tab:quora}.
Every 500 batch, we save the model which get the lower ELBO on the validation set.

\begin{table*}[!h]
 \begin{minipage}[h]{0.45\textwidth}
  \centering
  \begin{tabular}{c|c} \hline
    Hyper-parameters & Value \\ \hline
      $\lambda^\text{KL}_{sem}  $ & 1.0 \\
      $\lambda^\text{KL}_{syn}$   & 1.0 \\
      $\lambda^\text{mul}_{sem}$   & 0.5 \\
      $\lambda^\text{mul}_{syn}$   & 0.5 \\
      $\lambda^\text{adv}_{sem}$  & 0.5 \\
      $\lambda^\text{adv}_{syn}$  & 0.5 \\
      $\lambda^\text{rec}_{sem}$   & 0.5 \\
      $\lambda^\text{rec}_{syn}$   & 0.5 \\ 
      batch size & 32 \\
      GRU Droprate  & 0.1 \\\hline
    \end{tabular}
    \caption{The hyper-parameters we used in PTB dataset}
    \label{tab:ptb}
  \end{minipage}
  \qquad 
  \begin{minipage}[h]{0.45\textwidth}
  \centering
    \begin{tabular}{c|c} \hline
    Hyper-parameters & Value \\ \hline
      $\lambda^\text{KL}_{sem}  $ & 1/3 \\
      $\lambda^\text{KL}_{syn}$   & 2/3 \\
      $\lambda^\text{mul}_{sem}$   & 5.0 \\
      $\lambda^\text{mul}_{syn}$   & 1.0 \\
      $\lambda^\text{adv}_{sem}$  & 0.5 \\
      $\lambda^\text{adv}_{syn}$  & 0.5 \\
      $\lambda^\text{rec}_{sem}$   & 1.0 \\
      $\lambda^\text{rec}_{syn}$   & 0.05 \\ 
      batch size & 50 \\
      GRU Droprate  & 0.3 \\ \hline
    \end{tabular}
    \caption{The hyper-parameters we used in Quora dataset.}
    \label{tab:quora}
  \end{minipage}
\end{table*}

\section{Appendix II: Syntax-Transfer Case Study}
We also provided a few examples in Table~\ref{tab:case} at the next page. We see in all cases that a plain VAE ``interpolates'' two sentences without the consideration of syntax and semantics, whereas our \method is able to transfer the syntax without changing the meaning much. In the first example, \method successfully transfer a ``subject-be-predicative'' sentence to a ``there is/are'' sentence. 
For the second example, the semantic reference has the same syntactic structure as the syntax reference, and as a result,  \method generates the same sentence as $\refsem$. For the last example, we transfer a ``there is/are`` sentence to a ``subject-be-predicative`` sentence, and our \method is also able to transfer the syntax.

\begin{table*}[b]
    \footnotesize
    \centering
    \resizebox{.9\textwidth}{!}{
    \begin{tabular}{l|rl} \hline 
    \multicolumn{1}{c|}
    {Semantic and Syntax Providers} & \multicolumn{1}{c}{Sytnax-Controlled Output} \\ \hline \hline
    \begin{tabular}[c]{rl}
    $\textbf{Ref}_\textbf{syn}$: & There is an apple on the table . \\
    $\textbf{Ref}_\textbf{sem}$: & The airplane is in the sky .  
    \end{tabular} & 
    \begin{tabular}[c]{rl}
    \textbf{VAE}:& The man is in the kitchen .\\ 
    \textbf{\method}: & \parbox{6cm}{There is a airplane in the sky .}
    \end{tabular} \\ \hline
    \begin{tabular}[c]{rl}
     $\textbf{Ref}_\textbf{syn}$: & The shellfish was cooked in awork . \\
     $\textbf{Ref}_\textbf{sem}$: & The stadium was packed with people .
    \end{tabular} & 
    \begin{tabular}[c]{rl}
    \textbf{VAE}:& \parbox{5cm}{The man was filled with people .}\\ 
    \textbf{\method}:& \parbox{6cm}{The stadium was packed with people .}
    \end{tabular} \\ \hline
    \begin{tabular}[c]{rl}
    $\textbf{Ref}_\textbf{syn}$:& The child is playing in the garden . \\
    $\textbf{Ref}_\textbf{sem}$:& There is a dog behind the door .
    \end{tabular} & 
    \begin{tabular}[c]{rl}
    \textbf{VAE}:& There is a person in the garden .\\ 
    \textbf{\method}:& \parbox{6cm}{A dog is walking behind the door .}
    \end{tabular} \\ \hline
    \end{tabular}%
    }
    \caption{Case studies of syntax transfer generation.}
\label{tab:case}
\end{table*}